\documentclass[runningheads]{llncs}
\usepackage{graphicx}
\usepackage{amssymb}
\usepackage{amsmath}
\usepackage{stmaryrd}
\usepackage{multirow}
\usepackage{tabularx}
\usepackage{booktabs}
\usepackage{bbm}
\DeclareMathOperator*{\argmin}{arg\,min}
\begin{document}
\title{Scribble-based Domain Adaptation via Co-segmentation}
%
%
\author{Reuben Dorent\inst{1}, Samuel Joutard\inst{1}, Jonathan Shapey\inst{1,2,3}, Sotirios Bisdas \inst{3}, Neil Kitchen \inst{3}, Robert Bradford \inst{3}, Shakeel Saeed \inst{4}, Marc Modat\inst{1}, S\'ebastien Ourselin\inst{1} and Tom Vercauteren\inst{1}}

%
\authorrunning{R. Dorent et al.}
%
\institute{School of Biomedical Engineering and Imaging Sciences, King’s College London \email{reuben.dorent@kcl.ac.uk} \and
Wellcome/EPSRC Centre for Interventional and Surgical Sciences, University College London \and
National Hospital for Neurology and Neurosurgery, London \and
UCL Ear Institute, University College London}
\maketitle              
\begin{abstract}
Although deep convolutional networks have reached state-of-the-art performance in many medical image segmentation tasks, they have typically demonstrated poor generalisation capability. To be able to generalise from one domain (e.g. one imaging modality) to another, domain adaptation has to be performed. While supervised methods may lead to good performance, they require to fully annotate additional data which may not be an option in practice. In contrast, unsupervised methods don't need additional annotations but are usually unstable and hard to train. In this work, we propose a novel weakly-supervised method. Instead of requiring detailed but time-consuming annotations, scribbles on the target domain are used to perform domain adaptation. This paper introduces a new formulation of domain adaptation based on structured learning and co-segmentation. Our method is easy to train, thanks to the introduction of a regularised loss. The framework is validated on Vestibular Schwannoma segmentation (T1 to T2 scans). Our proposed method outperforms unsupervised approaches and achieves comparable performance to a fully-supervised approach. 
\end{abstract}
\section{Introduction}
Deep Neural Networks (DNNs) are achieving state-of-the-art performance for many medical image segmentation tasks. However, deep networks still lack in their generalisation capability when confronted with new datasets.

Domain adaptation (DA) approaches have been developed to ensure that networks trained on a source domain can be successfully used on a target domain. A first supervised solution consists of annotating (a sufficient number of) new images from the target domain and fine-tune a network initially trained on the source domain \cite{Supervised1,Supervised2}. Although easy to implement, stable during training, and achieving satisfying performance, such supervised techniques may not be a practical option given the time and expertise required to manually segment additional medical images. For this reason, unsupervised methods, based on self-supervised learning and adversarial learning, have been proposed. Self-supervised techniques~\cite{MauricioDA,PERONE} typically use pretext tasks to learn task-agnostic feature representations that are adapted to the target domain. Example of self-supervision includes optimising for prediction consistency across different strongly augmented versions of the same target data \cite{MauricioDA}. Although these techniques have shown promising results, they have only been tested on relatively similar source and target domain. Alternatively or concurrently, adversarial learning has been used to ensure that the learned feature representations are similar across the two domains via a discriminator network \cite{ADDA,DANN,MauricioDA,KostasDA,PnP-AdaNet,UnsupervisedAdver}. Relying on a complex and unstable adversarial optimisation procedure based on many heuristics, successfully training these models is particularly challenging and time-consuming. 
Moreover, they are often limited to 2D models due to high memory requirements.

In parallel, efforts have been done to help clinicians segment medical images more efficiently. In particular, semi-automated segmentation 
has been shown to be a reliable option 
\cite{DeepIGeoS}. Based on efficient user interactions such as scribbles, DNN predictions
are fine-tuned at an image-specific level \cite{BIFSEG}. Fine-tuning 
is performed for each new test image and is typically \emph{forgotten} on purpose afterwards as the image-specific nature implies a poorer generalisation capability. Looking beyond single images to streamline the annotation task, weakly-supervised methods based on scribbles have been introduced. Networks trained using scribbles are used to perform inference on unseen and unlabelled data. A standard modelling approach 
is to rely on Conditional Random Fields (CRFs) with DNN outputs being used as unary weights \cite{ScribbleSup,ScriMICCAI2,ScriMICCAI1}. The optimisation procedure typically alternates between proposing a one-hot crisp segmentation proposal extending from the scribbles (e.g. via a mean-field or graph-cut approach) and training the DNN with supervision provided by these proposals. A recent work \cite{RegularizedLoss} has shown that this two-step alternate optimisation can be efficiently approximated by a direct loss minimisation problem exploiting a regularised loss formulation.
\begin{figure}[tb!]
\begin{center}
\includegraphics[width=\linewidth]{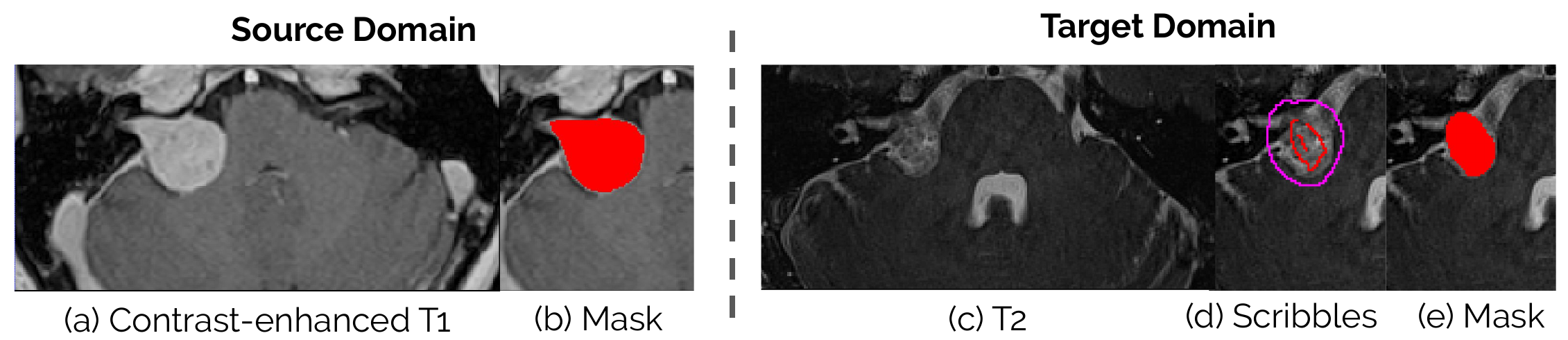}
\caption{Examples of Vestibular Schwanomma tumours. T1-c (a) and T2 (c) scans from the source and target domain are shown with their segmentation (b+e). Source masks (b) and target scribbles (e) are used at training stage.}
\label{example_scribbles}
\end{center}
\end{figure}

In this work, we propose a novel weakly-supervised domain adaptation method. The contributions of this work are four-fold.  First, we introduce a new formulation of domain adaptation as a co-segmentation problem. Secondly, we present a new structured learning approach to propagate information across domains. Thirdly, we show that alternating the proposal generation and network training can be approximated by directly minimising a regularised loss. Fourthly, we evaluate our framework on a challenging problem, unpaired cross-modality domain adaptation. Our method demonstrates the benefits of leveraging source data and obtained similar results compared to a fully-supervised approach.

\section{Conditional Random Fields for Structured Predictions}
In this section, we briefly present Conditional Random Fields (CRFs) for semantic segmentation and define some notations used in the remainder of this work. CRFs have been commonly used in image segmentation for their ability to produce structured predictions. 

Let $\mathbf{Y}$ be the random variable representing the overall label assignment, i.e. the segmentation, of a random image $\mathbf{Y}\in\mathbb{R}^{N}$, where $N$ is the number of voxels. For each voxel $k$, $Y_k$ is an element of the set of $C$ possible classes $\mathcal{L}=\{l_{1},\dots ,l_{C}\}$.
The general idea of CRFs is to model the pair $(\mathbf{X},\mathbf{Y})$ as a graph where the nodes (i.e. voxels) are associated with voxel-wise labels and the edges are associated with the similarity between the voxels.
Specifically, a CRF is characterised by a Gibbs distribution $P(\mathbf{Y}=\hat{\mathbf{y}}|\mathbf{X})\propto \exp(-E_{I}(\hat{\mathbf{y}}|\mathbf{X}))$. Here $E_{I}(\hat{\mathbf{y}}|\mathbf{X})$ is the Gibbs energy and represents the cost associated to the label configuration $\hat{\mathbf{y}}\in \mathcal{L}^{N}$. Given an observed image $\mathbf{x}$, the optimal segmentation  $\mathbf{\hat{y}^{*}}$  minimises the assignment cost. In the fully-connected pairwise CRF model, the problem is defined as:
\begin{equation}
    \label{eq1}
    \mathbf{\hat{y}^{*}} = \argmin_{\mathbf{\hat{y}}\in \mathcal{L}^{N}} \big\{E_{I}\left(\hat{\mathbf{y}}|\mathbf{x} \right)  = \sum_{k\in \llbracket 1;N \rrbracket}  \psi_{u}\left(\hat{y_k}|\mathbf{x} \right) +\sum_{k,l \in \llbracket 1;N \rrbracket}\psi_{p}\left(\hat{y_k},\hat{y_l}|\mathbf{x}\right) \big\}
\end{equation} 
where $\psi_{u}\left(\hat{y_k}|\mathbf{x}\right)$ and $\psi_{p}\left(\hat{y_k},\hat{y_l}|\mathbf{x}\right)$ are the unary and pairwise potentials.

Partial annotations, such as scribbles, provide known class values for a subset of voxels. Since each voxel depends on its neighbours, the sparse annotation information can be propagated within the image. Let $\mathbf{y}=(y_i)_{i\in\Omega_{a}}\in \mathcal{L}^{|\Omega_{a}|}$ be a partial annotation, where $\Omega_{a}$ is the set of annotated voxels (i.e. the scribbles). The optimisation problem then becomes a constrained one:
\begin{equation}
\label{eq:constrained}
\begin{split}
&\mathbf{\hat{y}^{*}} = \argmin_{\mathbf{\hat{y}}\in \mathcal{L}^{N}} 
\big\{ \sum_{k\in \llbracket 1;N \rrbracket}  \psi_{u}\left(\hat{y_k}|\mathbf{x} \right) +
\sum_{k,l \in \llbracket 1;N \rrbracket}\psi_{p}\left(\hat{y_k},\hat{y_l}|\mathbf{x}\right) \big\}\\
&\text { subject to }: \quad \forall k \in \Omega_{a}, \ \hat{y}_{k}=y_{k}
\end{split}
\end{equation}
The problem is typically solved by graph-cut \cite{GraphCut} for submodular problems or mean-field inference \cite{DenseCRF,ParallelMeanField} for the general case.

Recent works have proposed to combine the strengths of deep learning and structured learning via CRFs \cite{BIFSEG,ScribbleSup,ScriMICCAI2,ScriMICCAI1}. The common idea consists in defining the unary potentials $\psi_u$ with a neural network $f_{\theta}$ parameterised by the weights $\theta$. Existing methods typically alternate between proposal generation, i.e. solving~\eqref{eq:constrained}, and network parameters learning with supervision from these proposals. Recently this alternate optimisation has been replaced by a direct optimisation via a regularised loss \cite{RegularizedLoss}, thereby reducing the optimisation complexity, the computational cost during training and at inference, and avoiding learning from synthetically generated labels. The formulation in \cite{RegularizedLoss} reads:
\begin{equation} 
\label{eq:regularised}
\argmin_{\theta} \big\{\sum_{k\in\Omega_a} H(y_k,p_k) + R(\mathbf{p_{\theta}}) \big\}
\end{equation}
where $\mathbf{p_{\theta}}=f_{\theta}(\mathbf{x})$ is the softmax output of the network, $H$ is the cross-entropy and $R$ is a regularisation term that encourages spatial and image intensity consistency. In the next section, we provide more details about this regularisation term and we show how we adapt it for domain adaptation purposes.

\begin{figure}[tb!]
\begin{center}
\includegraphics[width=0.9\linewidth]{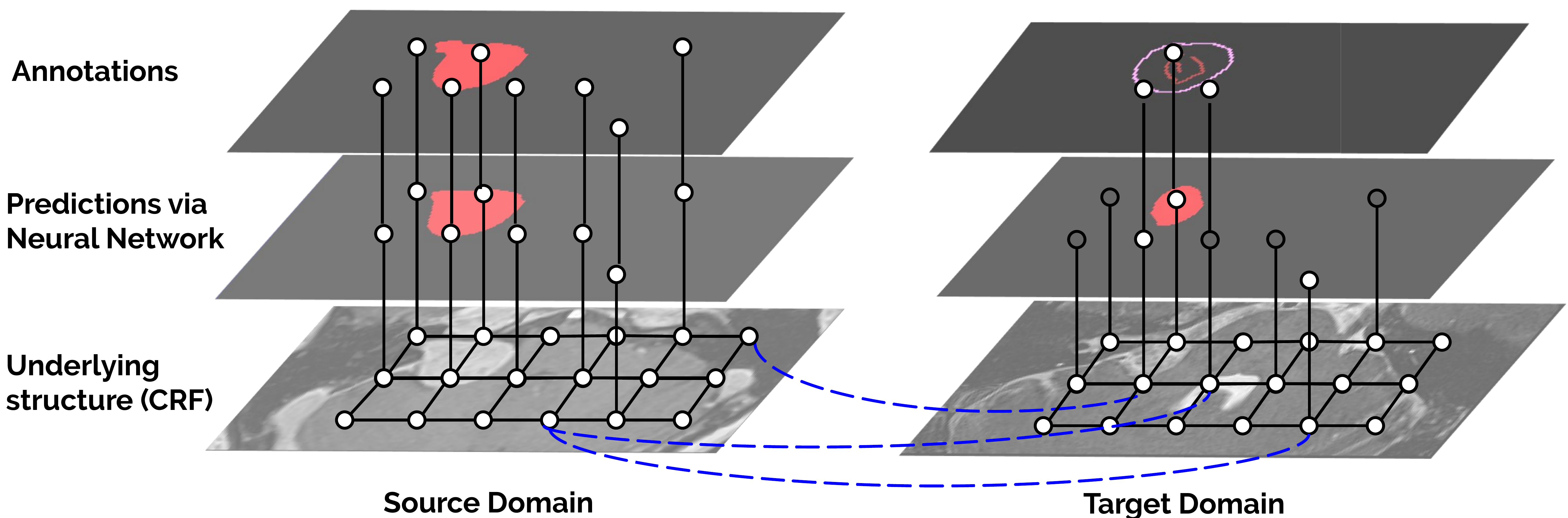}
\caption{Overview of the proposed graphical model. Each image voxel is a node. Annotations impose constraints on the predictions using a neural network $f_{\theta}$.  All the nodes are connected together within each image (image-specific CRF) and between images (domain adaptation; blue dashed lines). Only a few of these connections are shown. Although only two images are represented, all the images are connected to each other within and between domains. }
\label{graphical_model}
\end{center}
\end{figure}

\section{The Scribble Domain Adaptation Model}
\subsubsection{Group co-segmentation formulation}
In weakly-supervised domain adaptation, we are given a source domain $\mathcal{D}_{s}=\{ (\mathbf{x}_i^s,\mathbf{y}_i^s) \} $ of $n_s$ fully-labelled samples and a target domain $\mathcal{D}_{t}=\{ (\mathbf{x}_i^t,\mathbf{y}_i^t) \} $ of $n_t$ partially annotated samples. We denote $\Omega_{i,a}$ the set of annotated voxels for each image 
$\mathbf{x}_i^s$ ($\Omega_{i,a}$ representing the entire image) or $\mathbf{x}_i^t$ ($\Omega_{i,a}$ representing scribbles).
Figure~\ref{example_scribbles} shows an example of scribbles used in this work. 

The overall objective is to predict accurate segmentation for the target data using a neural network $f_{\theta}$. Since the annotations are partial on the target domain, we use a graphical model (a CRF) to include prior contextual information and perform structured predictions.
This allows for propagating the partial annotation information within a particular image. Given data from the target domain, we aim to minimise each image-specific Gibbs energy $E_{I}$, as defined in \eqref{eq1}.
However, this basic formulation does not include the other important source of information we have access to: The fully-annotated data from the source domain. 

Inspired by co-segmentation \cite{Coseg_Bach,Coseg_Hoch}, we extend the image-specific CRF to a dataset-level CRF. Specifically, in addition to including typical image-specific pairwise potentials, each node (i.e. voxel) of each image is connected to every nodes of every other images, as shown in Figure~\ref{graphical_model}. The annotation information is then propagated between images, including from the fully-annotated images to the partially-annotated images. Consequently, knowledge is transferred from the source domain to the target domain, i.e. domain adaptation is performed. For this reason, we denote $E_{DA}$ the proposed energy 
term
associated to pairs of images. Our proposed optimisation problem can be defined as follows:
\begin{equation}
\label{eq:final_formulation}
\begin{aligned}
&\argmin_{\theta,(\mathbf{\hat{y_i}})_i\in \mathcal{L}^{S\times N}} \Big\{ \sum_{i\in \llbracket 1;S \rrbracket} \big( E_{I}(\mathbf{\hat{y_i}|\mathbf{x_i}}) + \sum_{\substack{j\in \llbracket 1;S \rrbracket, 
j\neq i}} E_{DA}(\mathbf{\hat{y_i},\hat{y_j}|\mathbf{x_i},\mathbf{x_j}}) \big)  \Big\} \\
&\text { subject to }:  \quad \forall i\in \llbracket 1;S \rrbracket, \ \forall k\in \Omega_{a}, \  \hat{y}_{i,k}=y_{i,k} 
\end{aligned}
\end{equation}
where  $S=n_s+n_t$ is the total number of scans, and indices $i,j$ correspond to image index while $k,l$ are voxels index. Note that the constraints impose that the proposals for the source training data are exactly their fully-annotated masks.

\subsubsection{Image-specific Gibbs energy}
We used a standard formulation of the unary and pairwise potentials for the image-specific energy $E_{I}$ defined in \eqref{eq:constrained}.
Similarly to \cite{ScribbleSup,BIFSEG,ScriMICCAI1}, the DNN $f_{\theta}$ is used to compute the unary potentials:
\begin{equation}
    \forall k\in \llbracket 1;N \rrbracket, \quad \psi_{u}(\hat{y}_{i,k}|\mathbf{x_i)}= -\log P_{\theta}(\hat{y}_{i,k}|\mathbf{x_i})
    =  H(\hat{y}_{i,k},p_{i,k;\theta}) 
\end{equation} 
where  $\mathbf{p_{i;\theta}}=f_{\theta}(\mathbf{x_i})$ is the probability given by softmax output of the DNN and $H$ the cross-entropy.
For the image-specific pairwise potentials, we follow the typical choice of using the Potts model and a bilateral filtering term:
$$
\forall k,l \in \llbracket 1;N \rrbracket, \ \psi_{p}\left(\hat{y}_{i,k},\hat{y}_{i,l}|\mathbf{x}\right)=\left[\hat{y}_{i,k} \neq \hat{y}_{i,l}\right] \exp \Big(-\frac{(x_{i,k}-x_{i,l})^{2}}{2 \sigma_{\alpha}^{2}} -\frac{d(x_{i,k},x_{i,l})^{2}}{2 \sigma_{\beta}^{2}} \Big)
$$
where $d(.,.)$ denotes the Euclidean distance between the pixel locations. By denoting $W_{i}$ the affinity matrix  of an image $\mathbf{x_i}$ \cite{RegularizedLoss}, $E_{i}$ can be relaxed as:
\begin{equation}
    \label{eq:R_I-image}
    \sum_{k,l \in \llbracket 1;N \rrbracket}\psi_{p}\left(\hat{y}_{i,k},\hat{y}_{i,l}|\mathbf{x_i}\right)=   \mathbf{\hat{y_i}}^T W_{i}(1-\mathbf{\hat{y_i}}) \triangleq  R_{I}(\mathbf{\mathbf{\hat{y_i}}})
\end{equation} 

\subsubsection{Domain adaptation Gibbs energy}
The DA Gibbs energy $E_{DA}$ only involves pairwise potential associated with voxels from different images. By minimising $E_{DA}$, we expect to assign similar labels to voxels with similar visual features representation across the datasets. Since the domains are shifted, the image intensity distributions are different between the two domains. Consequently the image intensity cannot be used as features. Instead, we propose to use the features extracted from the DNN.
Specifically, we used the output of the penultimate convolution, i.e. just before the softmax regression. The domain adaptation cost is then defined as:
\begin{equation}E_{DA}(\mathbf{\hat{y_i},\hat{y_j}|\mathbf{x_i},\mathbf{x_j}}) = \sum_{k,l\in \llbracket 1;N \rrbracket} \left[\hat{y}_{i,k} \neq \hat{y}_{j,l}\right] \exp \Big(-\frac{(g_{i,k}-g_{j,l})^{2}}{2 \sigma_{\gamma}^{2}} \Big) 
\end{equation}
where $\mathbf{g}_{i}=g_{\theta}(\mathbf{x}_{i})$. Note that the spatial position is not taken into account here. Again, the domain adaptation Gibbs energy can be relaxed as:
\begin{equation}
    \label{eq:R_da-image}
    E_{DA}(\mathbf{\hat{y_i},\hat{y_j}|\mathbf{x_i},\mathbf{x_j}})= [\mathbf{\hat{y_i}},\mathbf{\hat{y_j}}]^T W_{i-j}(1-[\mathbf{\hat{y_i}},\mathbf{\hat{y_j}}]) \triangleq R_{DA}(\mathbf{\hat{y_i}},\mathbf{\hat{y_j}};\theta)
\end{equation} 

\section{Optimization via a Regularised Loss}
In this section, we propose a method to optimise the parameters $\theta$ of the DNN. Similarly to \cite{RegularizedLoss}, we show that the optimization problem can be approximated with a regularised loss.
Let $\mathbf{p^t_{i;\theta}}=f_{\theta}(\mathbf{x^t_{i}})$ and $\mathbf{p^s_{i;\theta
}}=f_{\theta}(\mathbf{x^s_i})$ be the outputs of the network for a target domain image $\mathbf{x^t_i}$ and a source domain image $\mathbf{x^s_i}$.  We denote $H_{\Omega_{a}}(\mathbf{u},\mathbf{v})=\sum_{k\in \Omega_{a}}H(u_k,v_k)$.
By combining \eqref{eq:final_formulation}, \eqref{eq:R_I-image} and \eqref{eq:R_da-image}, the optimisation problem is defined as:
\begin{equation}
\label{eq:full_opt}
\begin{split}
&\underset{\theta,(\mathbf{\hat{y_i}})_{i\in \llbracket 1;n \rrbracket}}{\argmin } \Big\{ \sum_{i} \big( H_{\Omega} \left(\mathbf{\hat{y_i}^s}, \mathbf{p^{s}_{i;\theta}}\right) + R_{I}(\mathbf{\hat{y_i}^s})\big) \\
&
\hphantom{xxxxxxxxxxx}
+ \sum_{i}\big( H_{\Omega} \left(\mathbf{\hat{y_i}^t}, \mathbf{p^{t}_{i;\theta}}\right) + R_{I}(\mathbf{\hat{y_i}^t})\big)+\sum_{i,j}R_{DA}\big(\mathbf{\hat{y_i}},\mathbf{\hat{y_{j}}}; \theta \big)  \Big\} \\
&\text { subject to }:  \quad \forall i\in \llbracket 1;S \rrbracket, \ \forall k\in \Omega_{a}, \ \hat{y}_{i,k}=y_{i,k} 
\end{split}
\end{equation}
%
By adding a null negative entropy term $-\sum_i H\left(\mathbf{\hat{y_i}^t},\mathbf{\hat{y_i}^t}\right)=0$ and integrating the constraints directly in the formulation, \eqref{eq:full_opt} can be rewritten as:
\begin{equation}
\argmin_{\theta} \Big\{ \sum_{i,j} u(\mathbf{p_{i,\theta}}) +  \mathbbm{1}_{\mathbf{x_i} \in \mathcal{D}^t} \min_{\mathbf{\hat{y_{i}}^{t}}} \left\{KL(\mathbf{\hat{y}^{t}_{i}},\mathbf{p_{i;\theta}})+R(\mathbf{\hat{y_{i}}^{t}}, \mathbf{\hat{y_{j}}^{t}}) \right\} \Big\} 
\end{equation}
where $u(\mathbf{p_{i,\theta}})=H_{\Omega_{a,i}}(\mathbf{y_i},\mathbf{p_{i,\theta}})$, $R(\mathbf{\hat{y_{i}}^{t}}, \mathbf{\hat{y}^{t}_{j}})=R_{I}(\mathbf{\hat{y_i}^{t}})+R_{DA}(\mathbf{\hat{y_i}^{t}}, \mathbf{\hat{y_{j}}^{t}})$ and $KL$ denotes the Kullback–Leibler divergence. Given that full annotations are provided for the source domain, the inner minimisation with respect to the proposals, $\mathbf{\hat{y}^{s}_{i}}$, only relates to the target data $(\mathbf{x}^{t}_{i},\mathbf{y}^{t}_{i})$. 

The inner problem corresponds to minimising a divergence between the network output $\mathbf{p^{t}_{i;\theta}}$ and the proposal $\mathbf{\hat{y_i}^{t}}$ together with a regularisation term. This discrepancy is null if the proposal is equal to the network output. We thus expect the optimal proposal to be close to the network output, i.e. $\mathbf{\hat{y_i}^{t*}}\approx \mathbf{p^{t}_{i;\theta}}$. We assume that equality stands, which allows us to reformulate the problem as:
\begin{equation}
\begin{split}
\underset{\theta}{\arg \min } \Big\{ \mathcal{L}(\theta) = \sum_{\substack{(\mathbf{x_i,y_i})\\ (\mathbf{x_j,y_j})}}H_{\Omega_{a,i}} (\mathbf{y^t_i}, \mathbf{p_{i;\theta}})+  \mathbbm{1}_{\mathbf{x_i} \in \mathcal{D}^t}( R_{I}(\mathbf{p_{i;\theta}})+R_{DA}(\mathbf{p_{i;\theta}},\mathbf{p_{j;\theta}}))  \Big\} 
\end{split}
\end{equation}
The parameters $\theta$ are directly optimised via a stochastic gradient descent. The high-dimensional filtering method proposed by \cite{Lattice} is used to reduce the quadratic complexity of the computation of $R_{I}$ and $R_{DA}$ to a linear one.

\begin{figure}[tb!]
\begin{center}
\includegraphics[width=\linewidth]{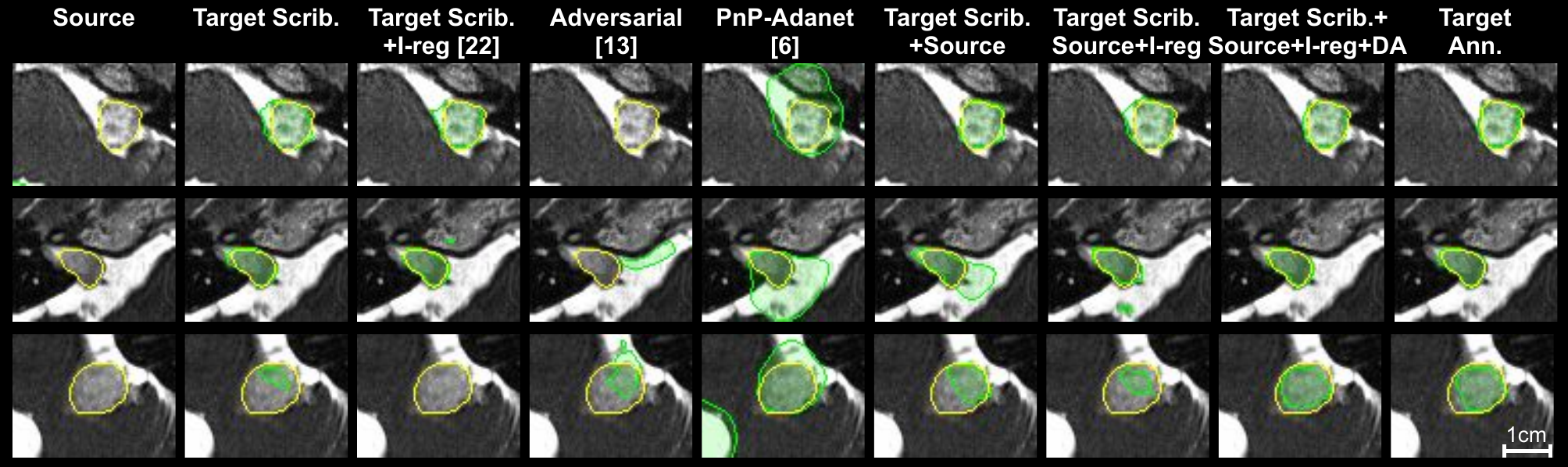}
\caption{Qualitative evaluation of different networks for Vestibular Schwannoma segmentation on T2 scans. Segmentation results (green curves) and the ground truth (yellow curves) are shown.}
\label{outputs}
\end{center}
\end{figure}

\section{Experiments}
\subsubsection{Experimental setup.} We conducted experiments on Vestibular Schwanomma (VS), a benign brain tumour arising from the vestibulocochlear  nerve, the main nerve connecting the brain and inner ear. Current MR protocols include contrast-enhanced T1-weighted (T1-c) and high-resolution T2 scans. T1-c are generally currently used for segmenting the tumour as offering a better contrast, see Figure~\ref{example_scribbles}. However, T2 imaging could be a reliable, safer and lower-cost alternative to T1-c \cite{VS_guideline,Jonathan}. 

In this work, we propose to segment VS images using T2 images only as input. The source domain data corresponds to 150 T1-c scans with the full set of annotations and the target domain training data corresponds to 30 T2 scans with scribble annotations only. Specifically, on average $1\%$ of the T2 scans and $7\%$ of the tumour has been annotated. 4 T2 scans and 20 T1 scans were used as validation set. For testing, 50 T2 scans (target domain) have been manually fully segmented. Images had an in-plane resolution of $0.4\times 0.4\times \text{mm}^2$, a slice thickness of $1.0-1.5\text{mm}$
and were cropped manually with a bounding box of size $100\times 50\times 50\text{mm}^3$, covering the full axial brain length as shown in Figures~\ref{example_scribbles}a,\ref{example_scribbles}c.


\subsubsection{Implementation details.} Our models were implemented in PyTorch using TorchIO \cite{fern2020torchio} \footnote{Code available at: \url{https://github.com/KCL-BMEIS/ScribbleDA}}. A 2.5D U-Net was used for all our experiments, similar to \cite{guotaiVS}. A PyTorch GPU implementation of the high-dimensional filtering \cite{SamuelJoutard} was employed. We used the Adam optimizer with weight decay $10^{-5}$.
At each iteration, two images from the source domain and two images from the target domain are randomly selected and fed to the network. The initial learning rate $5.10^{-4}$ was reduced by a factor of $5$ whenever the moving average of the validation loss has not improved in the last 5 epochs and training was stopped after no improvements in the last 10 epochs. Rotation, scaling and white noise augmentation were applied during training. 

Concerning the regularisation terms, a typical value of $\alpha$ was chosen ($15$). Similar results results were obtained for different values of $\beta$ ($\{0.5,0.05,0.005\}$), the ones reported correspond to $\beta=0.05$. In order to reduce the computational complexity, only two channels were used to compute the pairwise distance in the DA regularisation term. Specifically, at each training iteration, two channels were chosen randomly among the total number of channels (here 48). $\gamma$ was set up to $0.1$. Domain adaptation regularisation was introduced after a few epochs (70). Finally, we observed large improvements by using the Dice loss instead of the cross-entropy, thus we reported scores with the Dice loss. 

\subsubsection{Model Comparison} 
Firstly, we studied each component of our method independently. As a baseline, we trained a model on the target scribbles only (Target Scrib) and with the regularised loss \cite{RegularizedLoss} (Target Scrib+I-reg). Then the source data was used during training without (Target Scrib+I-reg+Source) and with (Target Scrib+I-reg+Source+DA) the cross-modality DA regularisation. Secondly, we compared our method with a fully-supervised approach trained using the same 30 T2 scans with the full set of annotations (Target Ann.). Thirdly, we compared our approach with two well-established unsupervised DA methods based on adversarial learning \cite{KostasDA} and designed specifically for cross-modality DA \cite{PnP-AdaNet}. Quantitative results are reported in Table~\ref{tab:WhiteMatterLesion:Dice} using the Dice and average symmetric surface distance (ASSD) between segmentation results and the ground truth. Examples of outputs are presented in Figure~\ref{outputs}.
\begin{table*}[t!]
    \caption{\textbf{Quantitative evaluation of different networks for Vestibular Schwannoma segmentation. I-reg: The image-specific regularised loss proposed by \cite{RegularizedLoss}. DA: Our proposed Domain Adaptation regularised loss.}}\label{tab:WhiteMatterLesion:Dice}
    \begin{tabularx}{\textwidth}{X *{5}{c}}
        \toprule
        &
        \multicolumn{2}{c}{\bf Test on Source }  & \multicolumn{2}{c}{\bf Tets on Target } \\
        \cmidrule(lr){2-3} \cmidrule(lr){4-5} 
        {\bf Method / Training}
        & Dice ($\%$) & ASSD ($\text{mm}^3$) & Dice ($\%$)  & ASSD ($\text{mm}^3$)   \\
        \midrule
        Source & \textbf{93.7 (3.3)}  & 0.3 (0.5)  & 28.2 (33.0)  & 13.8 (10.8) \\
        Target Scrib & 46.9 (33.8)  & 10.9 (10.9)  & 77.6 (17.9)  & 2.1 (3.0) \\
        Target Scrib+I-reg \cite{RegularizedLoss} & 58.4 (29.5)  & 9.0 (8.5)  & 76.9 (18.8)  & 1.4 (2.0) \\
        \midrule
        Source+Adversarial \cite{KostasDA} & 87.8 (8.9)  & 1.6 (1.5)  & 9.3 (18.9)  & 24.9 (16.2) \\
        PnP-Adanet \cite{PnP-AdaNet} & 79.3 (15.2)  & 3.4 (3.1)  & 27.3 (21.1)  & 13.3 (4.2) \\
        \midrule
        Target Scrib+Source & 92.4 (4.6)  & 0.4 (0.4)  & 75.1 (18.6)  & 2.7 (4.5) \\
        Target Scrib+Source+I-reg & 93.2 (3.7)  & 0.3 (0.5)  & 76.7 (17.9)  & 1.6 (2.5) \\
        Target Scrib+Source+I-reg+DA & 93.3 (4.0)  & \textbf{0.2 (0.2)}  & \textbf{83.4 (10.4)}  & \textbf{0.8 (0.8)} \\
        \midrule
        Target Ann. & 63.7 (33.9)  & 8.3 (11.2)  & 81.6 (13.1)  & 1.8 (2.8) \\
        \bottomrule
    \end{tabularx}
\end{table*}
\subsubsection{Results} Firstly, the ablation study shows that adding the cross DA regularisation brings significant improvements on the target domain compared to the other models trained using the target scribbles. Interestingly, including the source data during training only leads to improvements when the DA regularisation is employed. This shows the effectiveness of our DA method. Moreover, note that our technique didn't degrade the performance on the source domain. Secondly, our method obtained comparable performance to a fully-supervised model. Thus, scribble-based DA is a reliable option for performing supervised DA. Thirdly, both unsupervised methods failed on our problem. Since the inner brain and tumour appearance vary greatly between the contrast-enhanced T1 and T2 scans, our problem is too challenging for unsupervised approaches, highlighting the need for supervision.

\section{Conclusion}
This paper proposes a novel approach for weakly-supervised domain adaptation. Based on co-segmentation and structured learning, we introduced a new formulation for domain adaptation with scribbles. Our approach is mathematically grounded, easy to implement, new and relies on reasonable assumptions. We validated our method on challenging experiments: unpaired cross-modality brain lesion segmentation. Our model achieved comparable performance to a model trained on a fully-annotated data and outperformed existing unsupervised techniques. This work shows that scribbles is a reliable option for performing domain adaptation.

\subsubsection{Acknowledgement}
This work was supported by the Engineering and Physical Sciences Research Council (EPSRC) [NS/A000049/1] and Wellcome Trust [203148/Z/16/Z]. TV is supported by a Medtronic / Royal Academy of Engineering Research Chair [RCSRF1819\textbackslash7\textbackslash34].
%
%
%

\bibliographystyle{splncs04}
\bibliography{biblio}
\end{document}